\documentclass{article}
\usepackage{arxiv,times}

\usepackage{amsmath,amsfonts,bm}

\def\eqref#1{equation~\ref{#1}}

\def\1{\bm{1}}

\DeclareMathAlphabet{\mathsfit}{\encodingdefault}{\sfdefault}{m}{sl}
\SetMathAlphabet{\mathsfit}{bold}{\encodingdefault}{\sfdefault}{bx}{n}

\def\sR{{\mathbb{R}}}

\usepackage{hyperref}
\usepackage{url}
\usepackage{graphicx}
\usepackage{booktabs}
\usepackage{xcolor}

\title{Depth Estimators Are Implicit Neural Fields for 3D Scene Geometry Inpainting and Reconstruction}

\author{Yingzhao Jian, Zihao Lin \& Hehe Fan* \\
College of Artificial Intelligence, Zhejiang University. *Corresponding author. \\
\texttt{hehefan@zju.edu.cn}
}

\begin{document}

\maketitle

\begin{abstract}
The 3D geometry of real-world scene data is often incomplete. Mainstream methods use depth estimators to inpaint missing structure.
However, their prediction results can be inconsistent with observed geometry, or unreliable on out-of-distribution data. 
To solve these problems, we propose Neural Depth Field (NDF). 
Our key insight is that a depth estimator can also be a scene-level implicit field. 
As an estimator, it adapts to the target domain by learning observed depth data.
As an implicit field, it fits the existing geometry to maintain consistency.
Under this view, NDF addresses both problems through a single test-time optimization.
Experiments show that NDF produces high-fidelity and globally consistent geometry across diverse scene data, ranging from indoor scans to satellite imagery. It reduces cross-view inconsistency by 63.3\% and improves inpainting accuracy by 23.1\%,
achieving state-of-the-art performance in 3D scene geometry inpainting.
The code is available at: https://github.com/Shadow-Dream/Neural-Depth-Field.
\end{abstract}

\section{Introduction}

3D scene data~\citep{scannet2017,habitat2019,hu2022sensaturban} is crucial for visual understanding \citep{hong2023threedllm,majumdar2024openeqa}, spatial modeling \citep{bar2025navigationworldmodels,zheng2024occworld}, and embodied intelligence \citep{krantz2020vlnce,jian2025endowinggpt4}. These downstream tasks require complete and high-fidelity geometry observations, while real-world sensory data is often structurally incomplete. For instance, indoor scans suffer from missing surfaces in occluded regions \citep{kinectfusion2011,bundlefusion2017}; drone scans exhibit geometric gaps from constrained flight perspectives \citep{hu2022sensaturban,yin2024cus3d}; and satellite imagery contains data voids caused by clouds, shadows, or matching failures \citep{satnerf2022,skyfallgs2026}.

A prevalent approach to recovering continuous geometry is to use depth estimators. These models are trained on large-scale geometric data and can recover complete depth maps from partial observations \citep{depthanythingv2,lin2025depth,infinidepth2026}. However, two main difficulties remain. First, the completed region must remain consistent with the observed part. Second, certain data (e.g., satellite imagery) may be out of distribution for common depth estimators.

To solve these problems, we propose Neural Depth Field (NDF), which builds upon pretrained depth estimators. Our key insight is that a depth estimator can be treated as both a one-shot predictor and a scene-level implicit depth field. As a predictor, it learns domain-specific priors for depth inpainting by optimizing on existing depth observations. As an implicit depth field, it fits the observed region to maintain geometric consistency. Under this perspective, NDF solves the two aforementioned challenges through a single unified test-time optimization objective.

We evaluate our method across a spectrum of challenging real-world scenarios, ranging from city-scale satellite mapping to complex indoor scans. Quantitative and qualitative results demonstrate that NDF achieves both geometric accuracy and global structural consistency in 3D scene inpainting. Compared to current state-of-the-art methods, NDF reduces cross-view inconsistency by 63.3\% and improves inpainting accuracy by 23.1\%. It captures fine-grained structures in the inpainted regions, such as small streetlamps in satellite imagery and thin wires in indoor scans. Meanwhile, it preserves both local geometric coherence and global cross-view consistency in the existing parts.

\begin{figure}[!t]
    \centering
    \includegraphics[width=\textwidth]{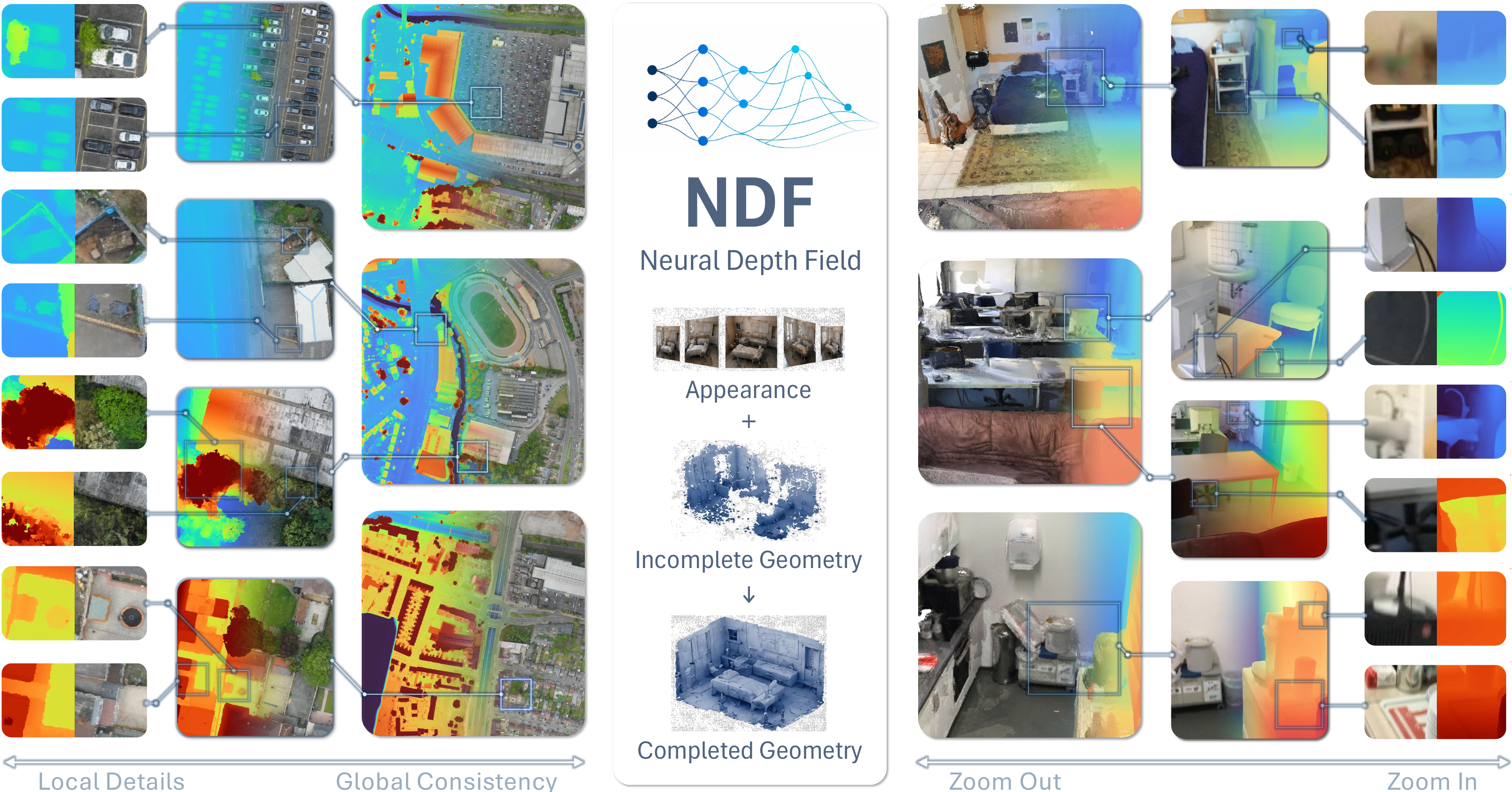}
    \caption{\textbf{Neural Depth Field (NDF)} is a test-time optimization approach designed for the scene geometry inpainting task. Given the scene's appearance (e.g., multi-view RGB images) and its partial geometry (e.g., incomplete depth maps) as input, NDF inpaints missing structures with high precision and global consistency. The results preserve fine-grained details such as chimneys in satellite imagery (left) and wires in indoor scans (right). Meanwhile, the inpainted regions blend seamlessly with the observed geometry and maintain coherence across overlapping inference windows.
    }
    \label{fig:ndf_overview}
\end{figure}

In summary, our contributions are threefold:

1. We are the first to systematically formulate depth estimators as implicit neural fields for unified 3D scene inpainting and reconstruction.

2. NDF achieves both state-of-the-art accuracy and consistency in 3D scene geometry inpainting on a variety of scene types, ranging from satellite imagery to indoor scans.

3. We provide a scalable pipeline for high-fidelity scene data inpainting and generation, together with processed scene datasets with completed geometry to support future research.

\section{Related Work}

\subsection{Depth Estimation and Completion}

Depth estimators learn geometric priors from large-scale data \citep{megadepth2018,hypersim2021,virtualkitti2,depthanything2024} and recover scene geometry from visual observations. Some methods directly predict relative depth from RGB images \citep{eigen2014depth,mvsnet2018,depthanythingv2,vggt2025}. However, they do not determine the metric scale, resulting in inconsistency between the prediction and the observed depth. 

Other methods use prompted depth \citep{ma2018sparse,nlspn2020,penet2021,promptdepthanything2025,infinidepth2026} to condition depth generation. Yet the conditioning signal is often sparse, so completed geometry can still show discontinuities along the inpainting boundaries.

NDF places observed depth directly in the optimization objective, so the pretrained geometric prior is retained while the field is forced to align with the metric scale of the current scene.

\subsection{Scene Reconstruction with Neural Fields}

Scene reconstruction recovers 3D structure from low-dimensional observations. Some methods directly optimize geometry from original measurements, such as multi-view images \citep{nerf2020,kerbl2023gaussian,satnerf2022,eogs2025,skysplat2025} or partial scans \citep{kinectfusion2011,elasticfusion2015,bundlefusion2017,badslam2019}. These methods have limited extrapolation ability for missing regions, for which no direct measurement is available.

Recent works further introduce depth estimators as additional supervision for neural-field optimization \citep{dsnerf2022,sparsenerf2023,monosdf2022,depthregularized3dgs2024,depthsplat2025}. However, this supervision is often indirect. The depth prediction and the optimized scene field remain separated and may become inconsistent.

NDF treats the depth estimator itself as the implicit geometry field. In this way, the depth prediction and scene field are inherently consistent.

\subsection{3D Scene Inpainting}

3D scene inpainting completes missing scene regions from incomplete observations. When inpainting geometry, most methods rely on pretrained priors for structural extrapolation \citep{nerfiller2024,infusion2024,3dgic2025,splatfill2025,inpaint360gs2026}. However, inpainting targets may differ from the data distribution on which the prior is learned. This gap makes direct prior extrapolation unreliable, as generated geometry can be mismatched to the current scene.

NDF fits the depth estimator on observed geometry from the target scene. This turns inpainting from direct pretrained prior extrapolation into scene-specific test-time adaptation.

\section{Method}

\begin{figure}[!t]
    \centering
    \includegraphics[width=\textwidth]{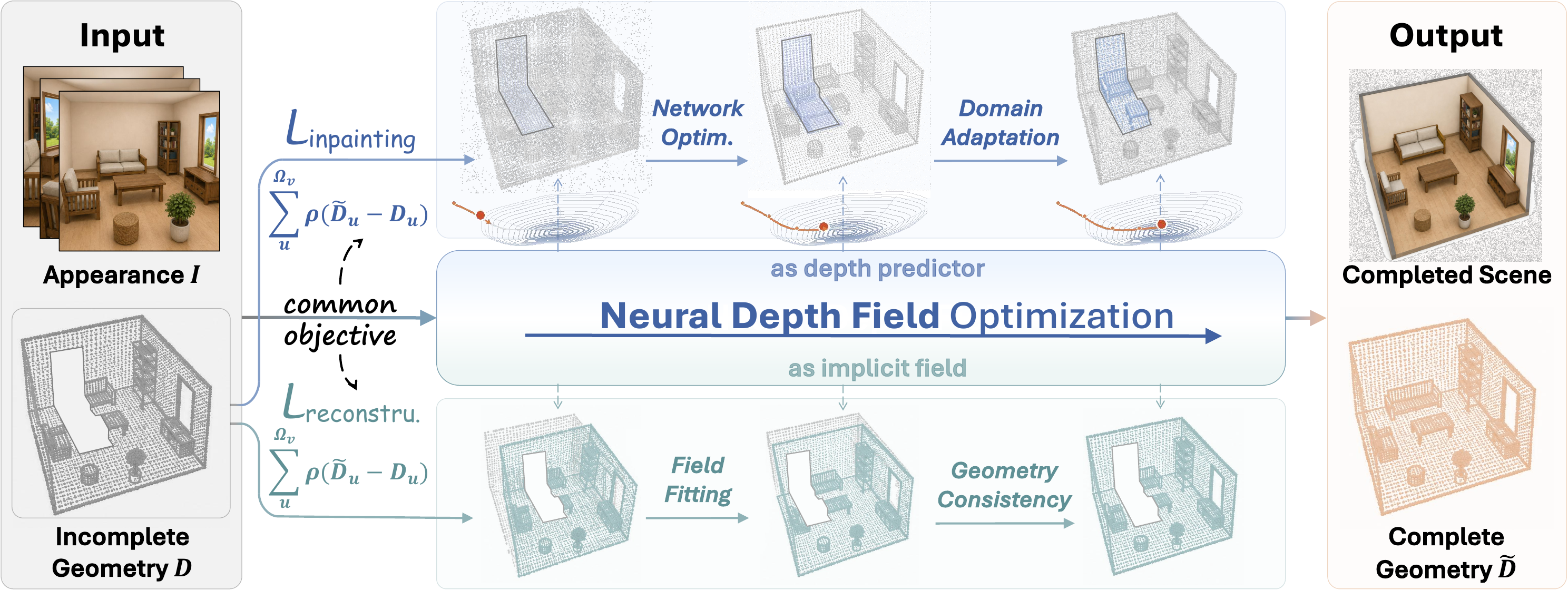}
    \caption{\textbf{Neural Depth Field Optimization.} NDF takes as input scene appearance \(I\) and incomplete geometry \(D\) with observed set \(\Omega_v\). At each optimization step, NDF renders scene geometry and predicts depth \(\tilde{D}\) conditioned on the appearance \(I\). As an \textbf{implicit neural field}, the rendered geometry \(\tilde{D}\) should fit the observed part \(D | _{\Omega_v}\). As a \textbf{depth estimator}, the prediction \(\tilde{D}\) should be supervised by the ground-truth depth \(D|_{\Omega_v}\) to adapt the prior distribution. Both objectives can be described by a pixel-wise discrepancy loss. By optimizing this shared loss, NDF samples inpainting from a domain-specific prior while maintaining consistency with the observed geometry.
    }
    \label{fig:ndf_optimization}
\end{figure}

\subsection{Problem Formulation}

We define a scene as an underlying world \(S\). An RGB-D image is an observation \(\mathcal{O}\) sampled from this scene. Given an observation parameter \(c\) that specifies the sampling configuration, such as position, scale, and viewing angle, the observation operator produces:
\begin{equation}
\mathcal{O}_c = (I_c, D_c, M_c) = \mathcal{R}(S,c).
\end{equation}

Each observation is defined on a pixel domain \(\Omega_c=\{1, \ldots, H\} \times \{1, \ldots, W\}\), where \(I_c: \Omega_c \to \mathbb{R}^3\) is the RGB image, \(D_c: \Omega_c \to \mathbb{R}\) is the incomplete scene depth, and \(M_c: \Omega_c \to \{0,1\}\) is the depth-validity mask.
When \(c\) is fixed, we write \(\mathcal{O}=(I,D,M)\) on \(\Omega\). Let \(u \in \Omega\) denote a pixel.
\begin{equation}
\Omega_v = \{u \in \Omega : M(u)=1\}, \qquad
\Omega_h = \Omega \setminus \Omega_v .
\end{equation}

Only \(D|_{\Omega_v}\) is observed. For an observation, the goal of depth inpainting is to find a depth estimator $f^*$, whose predicted depth $\tilde D$ best fits the complete depth observation $\hat D$. Without loss of generality, we assume that \(f\) is a monocular depth estimator to simplify the derivation:
\begin{equation}
\tilde D = f(I, D, M), \qquad f^* = \arg\min_f |\hat D-\tilde D|.
\end{equation}

\subsection{Neural Depth Field}

Let \(f_{\theta}\) be a mono-depth estimator with parameters \(\theta\). For the input image \(I\), it gives a one-shot depth prediction \(f_{\theta}(I)\). Since \(I\) is fixed for a given observation, \(\theta\) defines a scene-level implicit depth field \(F_{\theta}: \Omega \to \mathbb{R}\), where:
\begin{equation}
F_{\theta} = f_{\theta}(I)
\implies
F_{\theta}(u) = [f_{\theta}(I)](u).
\end{equation}

Therefore, the mono-depth model has two roles. As a predictor, \(f_\theta\) maps RGB to depth. As a field, \(F_{\theta}\) represents the depth surface of the current observation. Our objective is to minimize the gap between the complete and the predicted depth:
\begin{equation}
\theta^* =
\arg\min_\theta \mathbb{E}_{c\sim\mu}[\ell_c(\theta)]
=
\arg\min_\theta \mathbb{E}_{c\sim\mu}
\left[
\int_{\Omega_c}
\rho(F_{\theta,c}(u),\hat D_{c}(u))\,d\nu_c(u)
\right],
\end{equation}
where \(\theta^*\) is the optimization target, \(\mu\) is the distribution of \(c\), \(\ell_c\) is the loss term, \(\rho: \mathbb{R} \times \mathbb{R} \to \mathbb{R}_+\) is a pointwise loss, and \(\nu_c\) is the normalized counting measure on \(\Omega_c\). The loss can be further decomposed into an observed-depth fitting term and a hidden-depth term:
\begin{equation}
\ell(\theta)
=
\int_{\Omega}
M(u)\rho(F_{\theta}(u),D(u))\,d\nu(u)
+
\int_{\Omega}
(1-M(u))\rho(F_{\theta}(u),\hat D(u))\,d\nu(u).
\end{equation}

Since \(\hat D\) is unknown on \(\Omega_{c,h}\), we can approximate it using a neural network \(g_{\lambda}\) based on the observation \(D\) and another available prior \(\Pi_{g_{\lambda}}\), where \(\lambda\) denotes the parameters. The optimization target is:
\begin{equation}
\lambda^*
=
\arg\min_\lambda
\mathbb{E}_{c\sim\mu}
\left[
\int_{\Omega}
M_{c}(u)\rho(g_{\lambda}(u),D(u))\,d\nu(u)
+
\Pi_{g_{\lambda}}
\right].
\end{equation}

Notice that \(g_{\lambda}\) can be any optimizable neural network, including \(F_{\theta}\). In this case:
\begin{equation}
\ell(\theta)
\sim
\int_{\Omega}
M(u)\rho(F_{\theta}(u),D_c(u))\,d\nu_c(u)
+
\alpha\Pi_{F_{\theta}},
\end{equation}
where \(\alpha\) is a constant, and \(\Pi_{F_{\theta}}\) can be a pretrained depth prior, a smoothness regularizer, or any other constraint. Through the NDF parameterization, this single optimization objective unifies inpainting in the missing regions and reconstruction in the observed regions.

\section{Experiments}
\subsection{Satellite Height Map Inpainting}

\subsubsection{Setting}

\paragraph{Task.}
A satellite height map \(D_s\in\sR^{H\times W}\) represents the geometry of a large-scale RGB-D ortho-rectified satellite imagery scene \(S=(I_s,D_s,M_s)\), where \(I_s\in\sR^{H\times W\times 3}\) is the satellite RGB image and \(M_s\in\{0,1\}^{H\times W}\) is the height-validity mask. It supports downstream applications including robot navigation, city-scale reconstruction, and urban analysis. Yet the height map is often incomplete because of stereo matching failures or sensor artifacts. The task of satellite height-map inpainting is to complete the missing geometry in \(D_s | _{M_s = 0}\) while preserving the observed height values in \(D_s | _{M_s = 1}\).

Each depth-estimator observation is a local crop or frame indexed by \(c\), with RGB \(I_c\), height-derived depth \(D_c\), and a validity mask \(M_c\). For a satellite imagery scene, the observation parameter \(c\) is the coordinate and size of a local window. This task is inherently challenging for two reasons. First, the city-scale scene requires sliding-window inference. However, current methods struggle to remain consistent across windows. Second, satellite height maps are out of distribution for common depth estimators, which are primarily trained to predict depth from ground-view images.

\paragraph{Dataset.}
We use the Birmingham scene as our city-scale satellite benchmark. It contains a variety of complex urban structures, including buildings, riverbanks, and forests. The RGB canvas has resolution \(13673 \times 14004\), with a pixel size of \(0.1\) m. The scene contains \(177{,}875{,}345\) valid height pixels and \(13{,}680{,}747\) invalid height pixels. The valid pixels are further split into two disjoint subsets: \(176{,}140{,}546\) pixels are used for training and depth prompting, while \(1{,}734{,}799\) pixels are reserved for evaluation only. Details about data preprocessing can be found in Appendix~\ref{app:birmingham_rgb_inpainting}.

\paragraph{Metrics.}
We report absolute relative depth error (AbsRel), height mean absolute error (MAE, in meters), and \(\delta_{1+x}\), the fraction of evaluated pixels whose relative depth error is within \(x\). Both metric-depth and relative-depth inpainting performance are tested. For relative depth estimators that do not recover metric scale, we evaluate them using the corresponding scale-shift-invariant (SSI) metrics.

Evaluation is conducted on three types of regions. \textit{Heldout} measures the difference between predictions and ground-truth heights on the evaluation pixels, reflecting inpainting accuracy. \textit{Overlap} measures the discrepancy among predictions from overlapping windows in invalid regions, and \textit{Observed} measures regression accuracy on known heights; together, they reflect prediction consistency.

\subsubsection{Implementation}

We instantiate NDF with the pretrained InfiniDepth~\citep{infinidepth2026} model, a state-of-the-art monocular depth estimator. The NDF is optimized on randomly sampled windows of Birmingham training pixels. Each window uses \(50\) to \(2500\) prompt-depth points and \(100{,}000\) dense query points. The model is trained for \(10\) epochs with a batch size of \(16\), which takes about one hour on a single NVIDIA RTX PRO 6000 Blackwell GPU. We set the learning rate to \(2\times10^{-5}\) for the DINOv3 ViT-L/16 backbone and \(1\times10^{-4}\) for the remaining parameters.

During training, we use MAE as the reconstruction loss and additionally introduce an overlap-window loss, which encourages overlapping windows to predict the same height for the same pixels. To adapt the metric-depth model to the inpainting task, we further apply transferred-mask augmentation. Specifically, for each window, we randomly transfer invalid masks from other windows, and pixels within the transferred mask are excluded from the prompt depth points.

\subsubsection{Quantitative Result}

\begin{table}[!t]
\centering
\small
\caption{Comparison of metric-depth inpainting accuracy across different methods on the Birmingham satellite imagery dataset, together with the impact of key NDF components on inpainting performance. \textbf{Bold} and \underline{underline} denote the best and second-best results, respectively. The results show that NDF improves both the inpainting accuracy and global consistency. The ablation studies further demonstrate the effectiveness of unifying depth inpainting and reconstruction.}
\resizebox{\linewidth}{!}{%
\begin{tabular}{l|ccc|ccc|ccc}
\toprule
\multicolumn{1}{c}{} & \multicolumn{3}{|c}{Heldout} & \multicolumn{3}{|c}{Overlap} & \multicolumn{3}{|c}{Observed} \\
\cmidrule(lr){2-4}\cmidrule(lr){5-7}\cmidrule(lr){8-10}
Method & AbsRel $\downarrow$ & MAE $\downarrow$ & $\delta_{1.025}$ $\uparrow$& AbsRel $\downarrow$& MAE $\downarrow$& $\delta_{1.025}$ $\uparrow$& AbsRel $\downarrow$& MAE $\downarrow$& $\delta_{1.025}$ $\uparrow$\\
\midrule
\textit{Dummy Baseline} & 0.1756 & 2.0900 & 0.4113 & 0.1396 & 1.6752 & 0.4062 & 0.1750 & 2.1791 & 0.4087 \\
InfiniDepth~\citep{infinidepth2026} & 0.0432 & 0.4584 & 0.7917 & 0.0199 & 0.2397 & 0.8506 & 0.0201 & 0.2701 & 0.8557 \\
InfiniDepth NDF (ours) & \textbf{0.0244} & \textbf{0.2668} & \textbf{0.8644} & \textbf{0.0083} & \textbf{0.0880} & \underline{0.9393} & \textbf{0.0084} & \textbf{0.1204} & \underline{0.9396} \\
\midrule
w/o augmentation & \underline{0.0270} & \underline{0.2910} & \underline{0.8589} & \underline{0.0093} & \underline{0.0974} & \textbf{0.9424} & \underline{0.0085} & \underline{0.120} & \textbf{0.9425} \\
w/o pretrained prior & 0.0314 & 0.3657 & 0.8089 & 0.0129 & 0.1354 & 0.8947 & 0.0130 & 0.1905 & 0.8975 \\
w/o depth estimator & 0.1031 & 1.2384 & 0.5863 & 0.0333 & 0.4099 & 0.7518 & 0.0505 & 0.6975 & 0.7207 \\
\bottomrule
\end{tabular}
}
\label{tab:satellite_height_metric_depth}
\end{table}

We evaluate NDF on the Birmingham dataset using both metric-depth and relative-depth estimators. The dummy baseline is implemented using the Navier-Stokes method, which serves as a lower-bound reference. We further conduct ablation studies on the metric-depth inpainting task to analyze the contribution of each component. Specifically, we disable the training augmentation (i.e., mask transferring and overlap loss) to assess its gain, and use a randomly initialized depth head to simulate the absence of pretrained depth priors. Finally, we remove the RGB input to isolate the contribution of NDF’s depth-estimation capability to geometric reconstruction.

Tab.~\ref{tab:satellite_height_metric_depth} and Tab.~\ref{tab:satellite_height_pure_mono_ssi} summarize the metric-depth and relative-depth results of NDF on Birmingham data. In the held-out region, NDF improves inpainting accuracy by 9.2\% and 23.1\% for metric and relative depth, respectively. NDF also enhances prediction consistency, reducing the \textit{Overlap} MAE by 63.3\% and the \textit{Observed} MAE by 55.4\%. These results demonstrate the effectiveness of NDF in improving both the inpainting accuracy and global consistency of depth estimators.

Ablation studies further show that training augmentation reduces \textit{Heldout} AbsRel by 9.6\%. The depth-prediction task improves geometry reconstruction, reducing \textit{Observed} MAE by 82.7\%. The pretrained depth prior further improves inpainting and reconstruction accuracy by 6.9\% and 4.7\%, respectively. These results provide further insight that depth inpainting and geometry reconstruction are mutually beneficial.

\begin{table}[!t]
\centering
\small
\caption{Comparison of the relative-depth inpainting accuracy of different methods on the Birmingham satellite imagery dataset. \textbf{Bold} denotes the best result. The results show that NDF improves relative-depth inpainting performance in both missing and observed regions.}
\resizebox{\linewidth}{!}{%
\begin{tabular}{l|ccc|ccc|ccc}
\toprule
\multicolumn{1}{c}{} & \multicolumn{3}{|c}{Heldout - SSI} & \multicolumn{3}{|c}{Overlap - SSI} & \multicolumn{3}{|c}{Observed - SSI} \\
\cmidrule(lr){2-4}\cmidrule(lr){5-7}\cmidrule(lr){8-10}
Method & AbsRel $\downarrow$& MAE $\downarrow$& $\delta_{1.1}$ $\uparrow$& AbsRel $\downarrow$& MAE $\downarrow$& $\delta_{1.1}$ $\uparrow$& AbsRel $\downarrow$& MAE $\downarrow$& $\delta_{1.1}$ $\uparrow$\\
\midrule
\textit{Dummy Baseline} & 0.1778 & 2.3150 & 0.5520 & 0.1305 & 1.5301 & 0.5156 & 0.1746 & 2.4384 & 0.4811 \\
InfiniDepth~\citep{infinidepth2026} & 0.1851 & 2.5441 & 0.5613 & 0.1496 & 1.7138 & 0.4875 & 0.1522 & 2.1700 & 0.5428 \\
InfiniDepth NDF (ours) & \textbf{0.1206 } & \textbf{1.6527} & \textbf{0.6910} & \textbf{0.1077} & \textbf{1.2371} & \textbf{0.5999} & \textbf{0.1101} & \textbf{1.5434} & \textbf{0.6928} \\
\bottomrule
\end{tabular}
}
\label{tab:satellite_height_pure_mono_ssi}
\end{table}

\begin{figure}[!t]
    \centering
    \includegraphics[width=\textwidth]{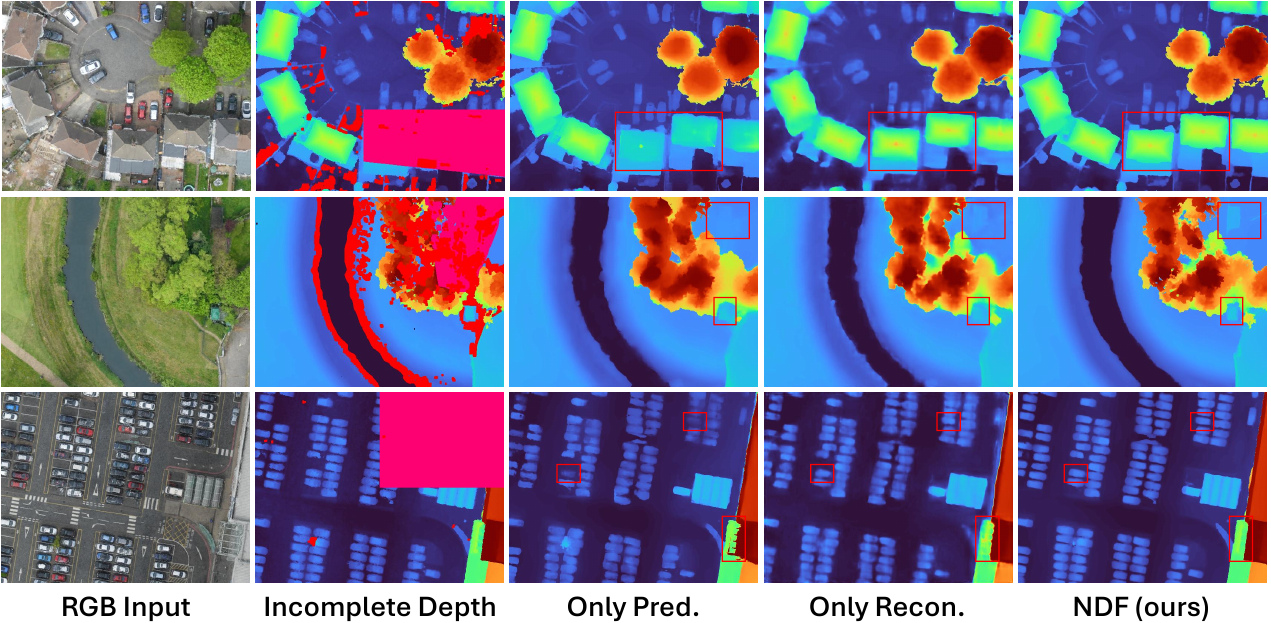}
    \caption{Visual comparison between the depth prediction results of different methods on the Birmingham satellite imagery. \textit{Only Pred./Recon.} denotes using only the pretrained depth predictor or only the geometry reconstruction module. In \textit{Incomplete Depth}, the pink and red regions indicate invalid pixels and held-out evaluation pixels, respectively. The pretrained depth estimator struggles to recover fine structures and plausible scale in the heldout regions. Reconstruction results are blurrier in detail and have limited extrapolation capability. NDF preserves global consistency while recovering high-fidelity details, such as walls and chimneys.}
    \label{fig:satellite_single}
\end{figure}

\subsubsection{Qualitative Result}

We visually compare the depth inpainting results of different methods on the Birmingham dataset. Specifically, the visualization samples cover diverse terrains such as urban areas, riverbanks, and forests. Within these samples, the heldout regions contain structures like vehicles, fences, and vegetation. These fine-grained details pose challenges to the inpainting accuracy of the models. Moreover, we visualize the results after stitching multiple overlapping prediction windows. The continuity between windows reflects the model’s ability to maintain global consistency.

\begin{figure}[!t]
    \centering
    \includegraphics[width=\textwidth]{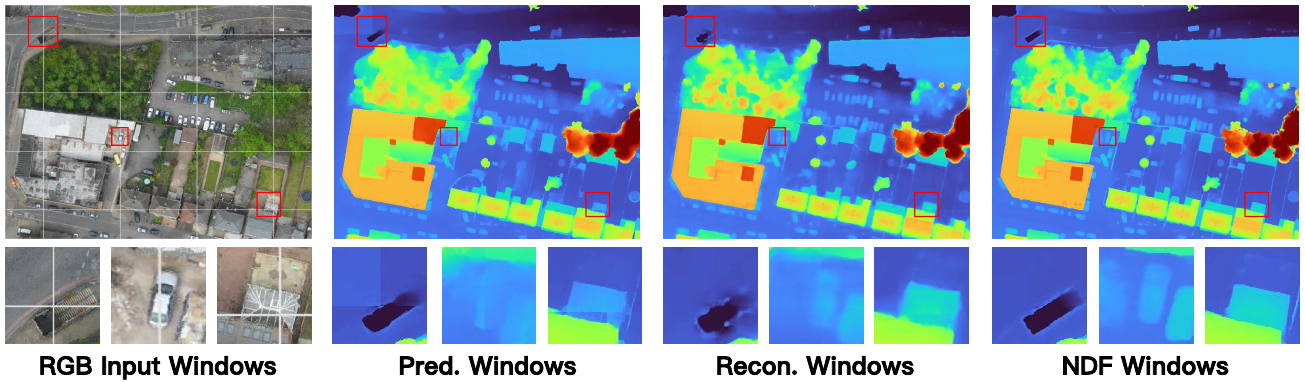}
    \caption{Comparison between the spliced depth prediction windows of different methods on the Birmingham satellite imagery. Directly using a pretrained depth estimator causes cross-window inconsistencies and hallucinated structures. Geometry reconstruction without depth prediction leads to blurry results. NDF ensures global consistency while preserving high-fidelity metric-scale details.}
    \label{fig:satellite_splice}
\end{figure}

Fig.~\ref{fig:satellite_single} and~\ref{fig:satellite_splice} show the single-window results and the stitched results from multiple windows, respectively. The pretrained depth estimator struggles to recover fine-grained details and plausible metric scale, which leads to cross-window inconsistencies and hallucinated structures. The reconstruction results are blurrier and exhibit limited extrapolation capability in heldout regions. NDF preserves global cross-view consistency while recovering high-fidelity details, demonstrating the effectiveness of unifying the inpainting and reconstruction objectives.

\subsection{Indoor Scan Depth Inpainting}
\providecommand{\textcolor}[2]{#2}
\providecommand{\ndfaudit}[1]{\textcolor{red}{#1}}

\subsubsection{Setting}

\paragraph{Task.}

An indoor scan represents the geometry of an indoor scene \(S\), where \(S\) can be an explicit or implicit geometry field. It supports downstream applications including robot vision and mixed reality. The indoor scans are often incomplete because of occlusions or sensor noise. The task of indoor scene inpainting is to complete the missing geometry while preserving the observed regions. 

For an indoor scene, the observation is usually an RGB-D rendering of the scene under a camera \(c\), with RGB \(I_c\in\sR^{H\times W\times 3}\), rendered depth \(D_c\in\sR^{H\times W}\), and a validity mask \(M_c\in\{0,1\}^{H\times W}\). When the camera intrinsics are predefined, \(c\) can be denoted by the extrinsics \([R_c, t_c]\), where \(R_c\) and \(t_c\) are the rotation and translation of the camera, respectively. As overlapping frames are related through perspective projection and occlusion, keeping the completion consistent across them is challenging. Adverse capture conditions—such as dim lighting, motion blur, and noisy odometry—further compound this difficulty.

\paragraph{Dataset.}

We use the ScanNet dataset \citep{scannet2017} to evaluate the performance of NDF on indoor scenes. 
ScanNet contains a variety of scene videos, including offices, living rooms, and kitchens. Each video frame provides an estimated camera pose, an RGB image, and an incomplete depth map. Depth incompleteness primarily arises from sensing failures or transparent surfaces. The data also includes artifacts such as pose inaccuracies, illumination variations, and motion blur.

For efficient evaluation, we sample 20 representative scenes with 38,679 frames. We construct two types of evaluation masks on these frames: 3D masks cover 1\%–5\% of each scene's volume. Depth pixels within them are removed across all frames.
2D masks are applied to 10\% of the frames. They remove depth pixels in image-space regions, while the underlying geometry may remain visible from other overlapping views. We exclude 20 frames before and after each selected frame to prevent information leakage.

\paragraph{Metrics.}
Following~\citet{infinidepth2026,lin2025depth}, we report absolute relative error (AbsRel), mean absolute error (MAE, in meters), and \(\delta_{1+x}\), the fraction of evaluated pixels whose relative depth error is within \(x\). We test both metric-depth and relative-depth inpainting performance. Relative depth estimators are evaluated using the corresponding scale-shift-invariant (SSI) metrics.

We evaluate the model's performance on three distinct region types.
\textit{Heldout} and \textit{Overlap} regions correspond to the 3D and 2D masks. The results in these regions reflect the geometric inpainting accuracy and cross-view consistency, respectively. In addition, \textit{Observed} regions measure regression accuracy on observed depth. 

\subsubsection{Implementation}

We perform NDF optimization using the pretrained InfiniDepth model. For each scene, the NDF is optimized for 1 epoch with a batch size of 1, which takes about 5 minutes on an NVIDIA RTX 4090 GPU. We set the learning rate to 2e-6 for the DINO backbone and 1e-5 for the remaining part. 

We use MAE loss for depth supervision and apply overlap-window loss and transferred-mask augmentation. Inspired by a recent method in 3D reconstruction~\citep{depthregularized3dgs2024}, we further incorporate a surface normal loss. It aligns the surface normals of NDF with the predictions of a teacher model, which helps mitigate artifacts caused by scan noise and preserve structural details.

\subsubsection{Quantitative Result}

\begin{table}[!t]
\centering
\small
\caption{Comparison of metric-depth inpainting accuracy across different methods on the ScanNet dataset, together with the impact of key NDF components on inpainting performance. \textbf{Bold} and \underline{underline} denote the best and second-best results, respectively. The proposed NDF model improves both inpainting accuracy in heldout regions and cross-view consistency in overlap views. The ablation results also demonstrate the effectiveness of unifying depth inpainting and reconstruction.}
\resizebox{\textwidth}{!}{%
\begin{tabular}{l|ccc|ccc|ccc}
\toprule
\multicolumn{1}{c}{} & \multicolumn{3}{|c}{Heldout} & \multicolumn{3}{|c}{Overlap} & \multicolumn{3}{|c}{Observed} \\
\cmidrule(lr){2-4}\cmidrule(lr){5-7}\cmidrule(lr){8-10}
Method & AbsRel $\downarrow$& MAE $\downarrow$& $\delta_{1.01}$ $\uparrow$& AbsRel $\downarrow$& MAE $\downarrow$& $\delta_{1.01}$ $\uparrow$& AbsRel $\downarrow$& MAE $\downarrow$& $\delta_{1.01}$  $\uparrow$\\
\midrule
\textit{Dummy Baseline} & 0.1303 & 0.1677 & 0.2005 
& 0.0777 & 0.1175 & 0.2935 
& 0.0783 & 0.1185 & 0.2907 \\
InfiniDepth~\citep{infinidepth2026} & 0.0272 & 0.0325 & 0.7010 
& 0.0122 & 0.0194 & 0.8096 
& 0.0053 & 0.0087 & 0.9617 \\
InfiniDepth NDF (ours) & \textbf{0.0236} & \textbf{0.0292} & \textbf{0.7621}
& \textbf{0.0080} & \textbf{0.0131} & \textbf{0.8588} 
& \underline{0.0035} & \underline{0.0058} & \underline{0.9740} \\ \midrule
w/o augmentation & \underline{0.0271} & \underline{0.0313} & \underline{0.7415} 
& \underline{0.0102} & \underline{0.0158} & \underline{0.8369} 
& \textbf{0.0033} & \textbf{0.0057} & \textbf{0.9761} \\
w/o pretrained prior & 0.0652 & 0.0839 & 0.2696 
& 0.0305 & 0.0508 & 0.3307
& 0.0479 & 0.0673 & 0.3002 \\
w/o depth prompt & 0.0666 & 0.0898 & 0.2009 
& 0.0301 & 0.0514 & 0.3120 
& 0.0328 & 0.0557 & 0.2950 \\
w/o depth estimator & 0.1189 & 0.1444 & 0.3534 
& 0.0431 & 0.0676 & 0.4820 
& 0.0037 & 0.0064 & 0.9599 \\
\bottomrule
\end{tabular}
}
\label{tab:scannet_metric_depth}
\end{table}

\begin{table}[!t]
\centering
\small
\caption{Comparison of the relative‑depth inpainting accuracy of different methods on the ScanNet dataset, along with the influence of training augmentation on inpainting performance. \textbf{Bold} denotes the best result in each column. The proposed NDF framework improves both inpainting precision and consistency. Ablation results demonstrate that NDF remains effective even without dedicated training design, and the training augmentation further contributes to inpainting performance.}
\resizebox{\textwidth}{!}{%
\begin{tabular}{l|ccc|ccc|ccc}
\toprule
\multicolumn{1}{c}{} & \multicolumn{3}{|c}{Heldout - SSI} & \multicolumn{3}{|c}{Overlap - SSI} & \multicolumn{3}{|c}{Observed - SSI} \\
\cmidrule(lr){2-4}\cmidrule(lr){5-7}\cmidrule(lr){8-10}
Method 
& AbsRel $\downarrow$& MAE $\downarrow$& $\delta_{1.05}$ $\uparrow$ 
& AbsRel $\downarrow$& MAE $\downarrow$& $\delta_{1.05}$ $\uparrow$
& AbsRel $\downarrow$& MAE $\downarrow$& $\delta_{1.05}$ $\uparrow$ \\
\midrule
\textit{Dummy Baseline}
& 0.1303 & 0.1677  & 0.6253  
& 0.0777 & 0.1175  & 0.7588  
& 0.0783 & 0.1185  & 0.7563  \\
InfiniDepth~\citep{infinidepth2026}
& 0.0450 & 0.0555  & 0.6920  
& 0.0416 & 0.0590  & 0.7368  
& 0.0416 & 0.0593  & 0.7366  \\
InfiniDepth NDF (ours)
& \textbf{0.0422} & \textbf{0.0409}  & \textbf{0.8124} 
& \textbf{0.0178} & \textbf{0.0234}  & \textbf{0.9394}  
& \textbf{0.0177} & \textbf{0.0236}  & \textbf{0.9380} \\ \midrule
w/o augmentation 
& 0.0436 & 0.0419  & 0.8029  
& 0.0184 & 0.0243  & 0.9363  
& 0.0185 & 0.0245  & 0.9355  \\

\bottomrule
\end{tabular}
}
\label{tab:scannet_relative_depth}
\end{table}

We evaluate NDF on the ScanNet indoor dataset, covering both metric-depth and relative-depth inpainting tasks. To isolate the contribution of each NDF component, we conduct ablation studies. Specifically, we ablate the training augmentation by disabling mask transferring and the additional loss terms. Pretrained depth priors are removed by reinitializing the depth head. The depth prompt is ablated by switching to the relative‑depth estimator. Finally, we remove NDF’s depth‑estimation role by discarding the RGB input. This ablation turns the NDF into a pure geometry field, which assesses the synergy between the depth inpainting and reconstruction tasks.

Tab.~\ref{tab:scannet_metric_depth} and Tab.~\ref{tab:scannet_relative_depth} present the metric- and relative-depth inpainting results, respectively. In the \textit{Heldout} regions, NDF achieves a 13.2\% reduction in AbsRel for metric-depth and delivers a 17.4\% accuracy gain for relative-depth over the original depth estimator. For inpainting consistency, the proposed NDF optimization lowers the \textit{Overlap} MAE by 32.5\% (metric) and 60.3\% (relative), and cuts the \textit{Observed} MAE by 33.3\% (metric)  and 60.2\% (relative). These results validate that NDF consistently improves both local inpainting fidelity and global structural coherence in indoor reconstruction tasks.

According to the ablation results, training augmentation reduces \textit{Overlap} AbsRel by 21.6\% for metric-depth. Even without dedicated data augmentation and training objective designs, NDF still improves inpainting accuracy by 5.8\% (metric) and 16.0\% (relative). 
The pretrained depth prior, the metric-depth prompt, and the depth estimation task each contribute to both the inpainting and reconstruction accuracy. These results verify the effectiveness of NDF and reveal that the inpainting and reconstruction tasks are mutually reinforcing. 

\subsubsection{Qualitative Result}

\begin{figure}[!t]
    \centering
    \includegraphics[width=\textwidth]{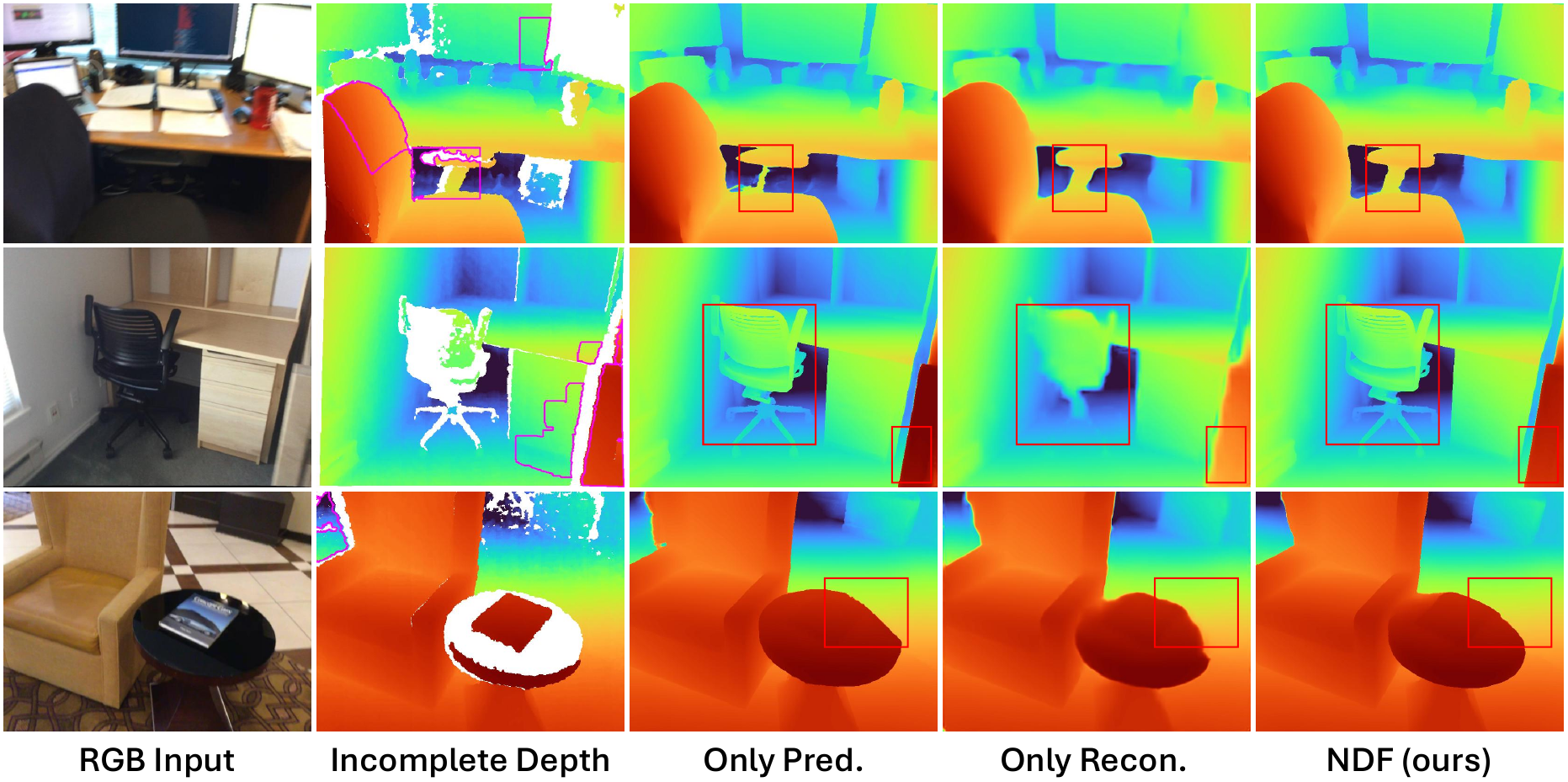}
    \caption{Visual comparison between the depth prediction results of different methods on ScanNet indoor scenes. \textit{Only Pred./Recon.} denotes using only the pretrained depth predictor or only the geometry reconstruction module. In \textit{Incomplete Depth}, white pixels indicate invalid depth, and pixels within the magenta borders are excluded from training and depth prompting. The pretrained depth estimator misses geometric structures in low‑light and reflective areas and struggles to recover metric scale in held‑out regions. The reconstructed geometry exhibits blurrier details. NDF unifies depth prediction and reconstruction, achieving noise robustness, metric-scale consistency, and high structural fidelity.}
    \label{fig:indoor}
\end{figure}

We conduct qualitative evaluation on the ScanNet dataset. Specifically, we visually compare NDF against the methods that perform depth prediction and reconstruction separately. The selected samples represent a variety of real-world scan scenarios, including low‑light regions, thin structures, motion blur, and reflective surfaces. These interfering factors pose significant challenges to both the inpainting fidelity and consistency of the compared methods.

The visualization results are shown in Fig.~\ref{fig:indoor}. In low‑light and reflective areas, the pretrained depth estimator fails to preserve structures like chair handles and table edges. The reconstructed geometry exhibits blurred boundaries and lacks the ability to extrapolate missing parts, such as the chair legs. Both the depth estimation and the reconstructed geometry suffer from limitations in recovering plausible metric scales in held‑out regions. NDF unifies depth inpainting and reconstruction, achieving noise robustness, metric‑scale consistency, and high structural fidelity.

\section{Conclusion}

We present Neural Depth Field (NDF), a test-time optimization framework for 3D scene geometry inpainting and reconstruction. Our key insight is that a pretrained depth estimator is simultaneously a one-shot predictor and a scene-level implicit neural field: as a depth predictor, it adapts its prior to the target domain under the supervision of observed depth; as an implicit field, it fits the observed geometry to maintain consistency. This dual view reduces inpainting and reconstruction to a single pixel-wise discrepancy objective. Experiments demonstrate the effectiveness of NDF on scenarios ranging from city-scale satellite imagery to complex indoor scans and show that it consistently enhances pretrained estimators in both inpainting accuracy and global consistency.

\bibliography{references}
\bibliographystyle{arxiv}

\newpage
\appendix

\section{Limitations and future work} 
First, NDF requires per-scene test-time optimization, taking minutes to an hour per scene; amortizing this cost through meta-learned initializations or parameter-efficient adaptation is a natural next step. Second, our current instantiation targets static scenes observed as depth and height maps; extending the formulation to dynamic scenes, richer geometric representations such as Gaussian splatting, and joint appearance–geometry completion remains an important direction. We leave these explorations to future work.

\section{Implementation Details}
\label{app:implementation_details}

\subsection{InfiniDepth Variants}

Our implementation is based on the official InfiniDepth repository\footnote{\url{https://github.com/zju3dv/InfiniDepth}}. The metric-depth model is the official InfiniDepth DepthSensor variant, it takes an RGB image and sparse metric depth prompts. The relative-depth model is the official InfiniDepth RGB-only variant, it takes only an RGB image and does not use depth prompts. Accordingly, the DepthSensor results are evaluated directly in metric space, whereas the RGB-only results use scale-shift-invariant (SSI) alignment.

\subsection{Model Architecture}

Both variants use a DINOv3 ViT-L/16 image backbone with a patch size of $16$. We extract intermediate features from transformer blocks $4$, $11$, $17$, and $23$. A shallow convolutional \textit{BasicEncoder} supplies $128$-dimensional low-level features at one-quarter input resolution. At an arbitrary query coordinate, bilinear sampling retrieves the DINOv3 and BasicEncoder features; the two features are concatenated and decoded by an implicit MLP with hidden dimensions $[1024,256,32]$ and an ELU output. The network predicts disparity, which is inverted to obtain depth.

DepthSensor additionally embeds each sparse depth value with a two-layer MLP and combines the depth and positional embeddings with the final-stage image tokens. Its prompt module contains four self-attention blocks with four heads. RGB-only shares the image backbone, BasicEncoder, and implicit decoder but disables this prompt module. This shared coordinate-query architecture allows both variants to produce predictions at resolutions different from the $672 \times 512$ encoder input.

\subsection{Prompt and Query Sampling}

All training frames are resized to $672 \times 512$ before feature extraction. For DepthSensor training, we randomly retain between $50$ and $2500$ valid depth-prompt points per frame. For validation and inference, we use $1500$ prompt points, or all available valid points when fewer than $1500$ exist. RGB-only receives no prompt depth during either training or inference. Pixels belonging to transferred invalid masks or held-out regions are removed from the prompt set, preventing target leakage.

We train the implicit decoder using sampled-query supervision rather than materializing the full output grid. Each training frame supplies $100{,}000$ query coordinates sampled from supervised pixels. At inference time, the complete output grid is queried in chunks to control memory use; this batching changes only memory consumption and not the predicted field. For Birmingham, metric heights are converted to local camera depth before prompting and converted back to height after prediction. For ScanNet, the rendered metric depth is used directly. The observed-depth MAE, transferred-mask augmentation, overlap consistency, and surface-normal regularization described in the main paper are applied on top of this common InfiniDepth parameterization.

\section{Birmingham RGB Inpainting}
\label{app:birmingham_rgb_inpainting}

\subsection{Setting}

The Birmingham satellite scene contains missing regions in both RGB and height. Before height-map inpainting, we complete the RGB canvas so that the depth estimator receives a dense visual observation. The RGB inpainting stage is designed for native-resolution city-scale completion, where local image quality and global writeback consistency must both be preserved.

We evaluate RGB completion across four representative region types: land interior holes, river continuity regions, sea or coastline regions, and estuary or river-mouth regions. These categories cover the dominant failure modes in the Birmingham scene: missing urban texture, discontinuous waterways, coastline drift, and large-region color or structure mismatch.

\subsection{Inpainting Agent}

We use a typed inpainting agent built on Nano Banana 2 as the image generation backbone and GPT-5.5 as the reviewer. The target region is first covered by a DFS-style overlapping-window plan. Each exported window is assigned a semantic type, and each type is dispatched to its own prompt template and reviewer configuration. This type-aware design lets the pipeline use different instructions for land, river, sea, coastline, and estuary regions.

The reviewer screens generated candidates before writeback. It rejects outputs with mask leakage, remaining unfilled pixels, broken shoreline or river continuity, excessive color drift, or edits that alter valid context outside the target mask. The reviewer configuration is tuned through few-shot self-iteration on a small manually labeled set of success and failure cases.

\subsection{Implementation Details}

The Birmingham RGB canvas has a native resolution of \(13673 \times 14004\), corresponding to approximately \(0.1\) m ground sampling distance. For the formal Stage-2 RGB completion run, the DFS planner produces \(588\) typed windows. After reuse and no-op pruning, \(519\) windows require actual generation.

Across the formal mainline runs, Nano Banana 2 evaluates \(968\) candidate images for the accepted \(519\)-window solution, with \(963\) realized upstream generation attempts and \(81\) upstream failures. The accumulated active generation time from recorded per-attempt latencies is about \(16.6\) hours, with an average of \(61.9\) seconds per candidate. The full mainline campaign spans about \(122.7\) wall-clock hours from the first formal run to the last archived center-only pass.

The GPT-5.5 reviewer is invoked \(882\) times in the formal pipeline, keeping \(521\) candidates and rejecting \(361\). The production reviewer logs do not store exact API usage, so we estimate the reviewer cost conservatively at roughly one million tokens in total.

\subsection{Model Selection}

We compare four upstream image-editing backbones for typed Stage-2 RGB completion: Nano Banana, GPT Image 2, Nano Banana Pro, and Nano Banana 2. The comparison uses a fixed \(16\)-window benchmark covering land, sea, and estuary cases.

\begin{figure}[!t]
    \centering
    \includegraphics[width=\textwidth]{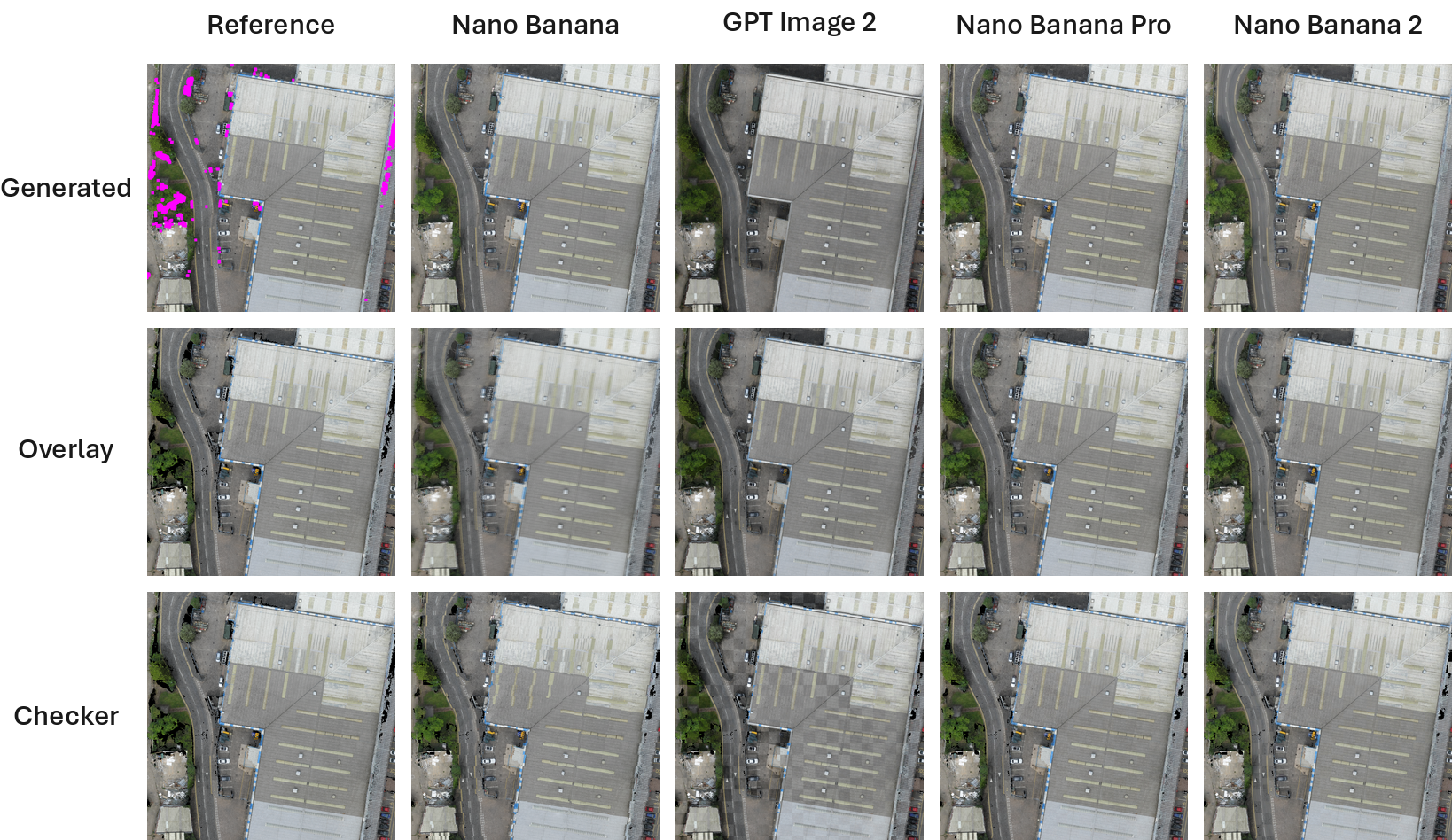}
    \caption{Comparison of RGB inpainting backbones on a land-region hole. Columns show the reference, Nano Banana, GPT Image 2, Nano Banana Pro, and Nano Banana 2; rows show the generated result, an overlay with the original context, and a checkerboard comparison. Nano Banana produces a visible global misalignment, whereas Nano Banana 2 preserves the surrounding urban registration and removes the mask completely.}
    \label{fig:app_rgb_model_land_alignment}
\end{figure}

\begin{figure}[!t]
    \centering
    \includegraphics[width=\textwidth]{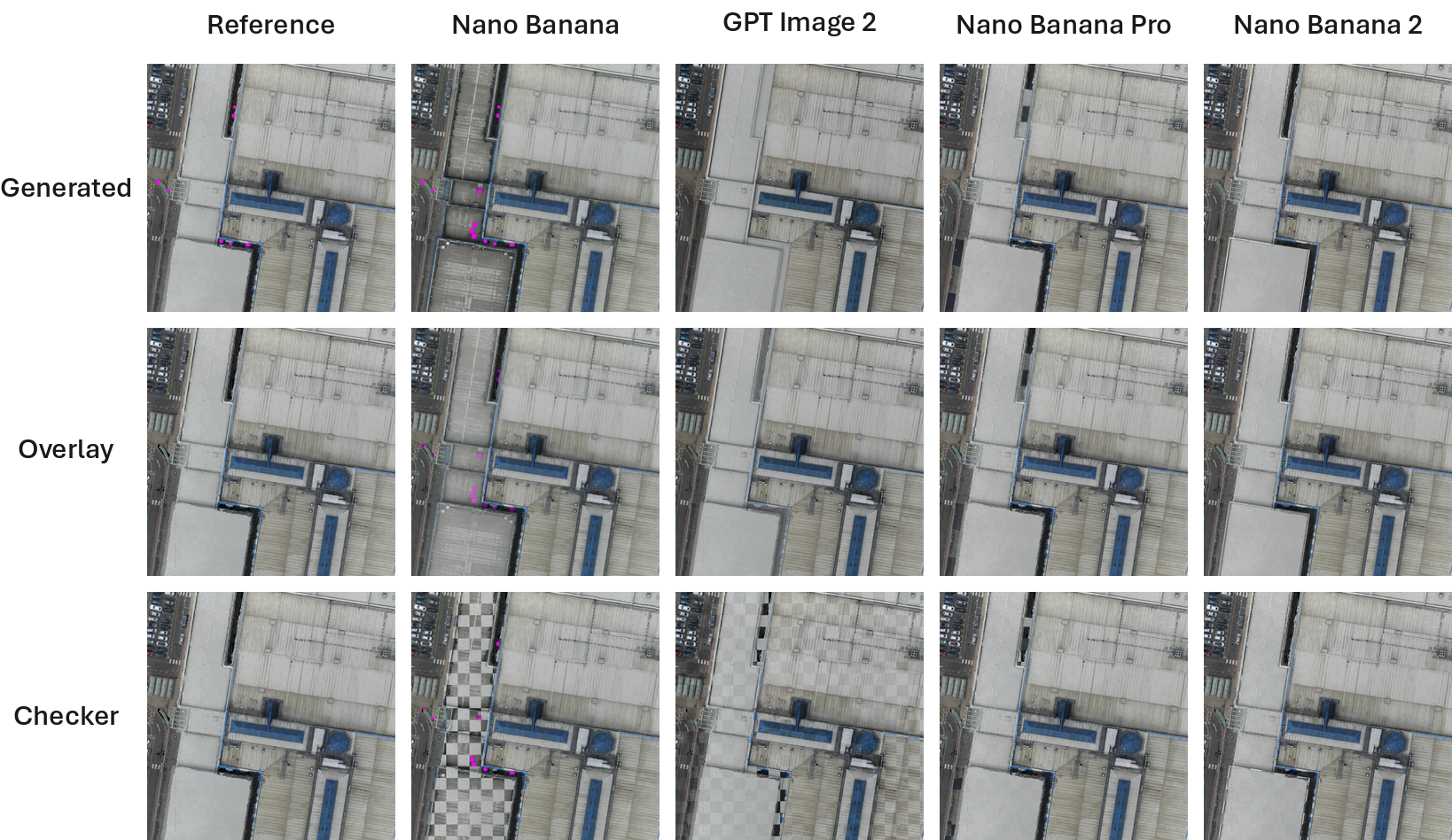}
    \caption{Comparison of RGB inpainting backbones on a small land-region hole. Nano Banana leaves part of the masked region unfilled, while GPT Image 2 changes the global color distribution. Nano Banana 2 completes the target region while better preserving the appearance of the valid context.}
    \label{fig:app_rgb_model_land_fill}
\end{figure}

\begin{figure}[!t]
    \centering
    \includegraphics[width=\textwidth]{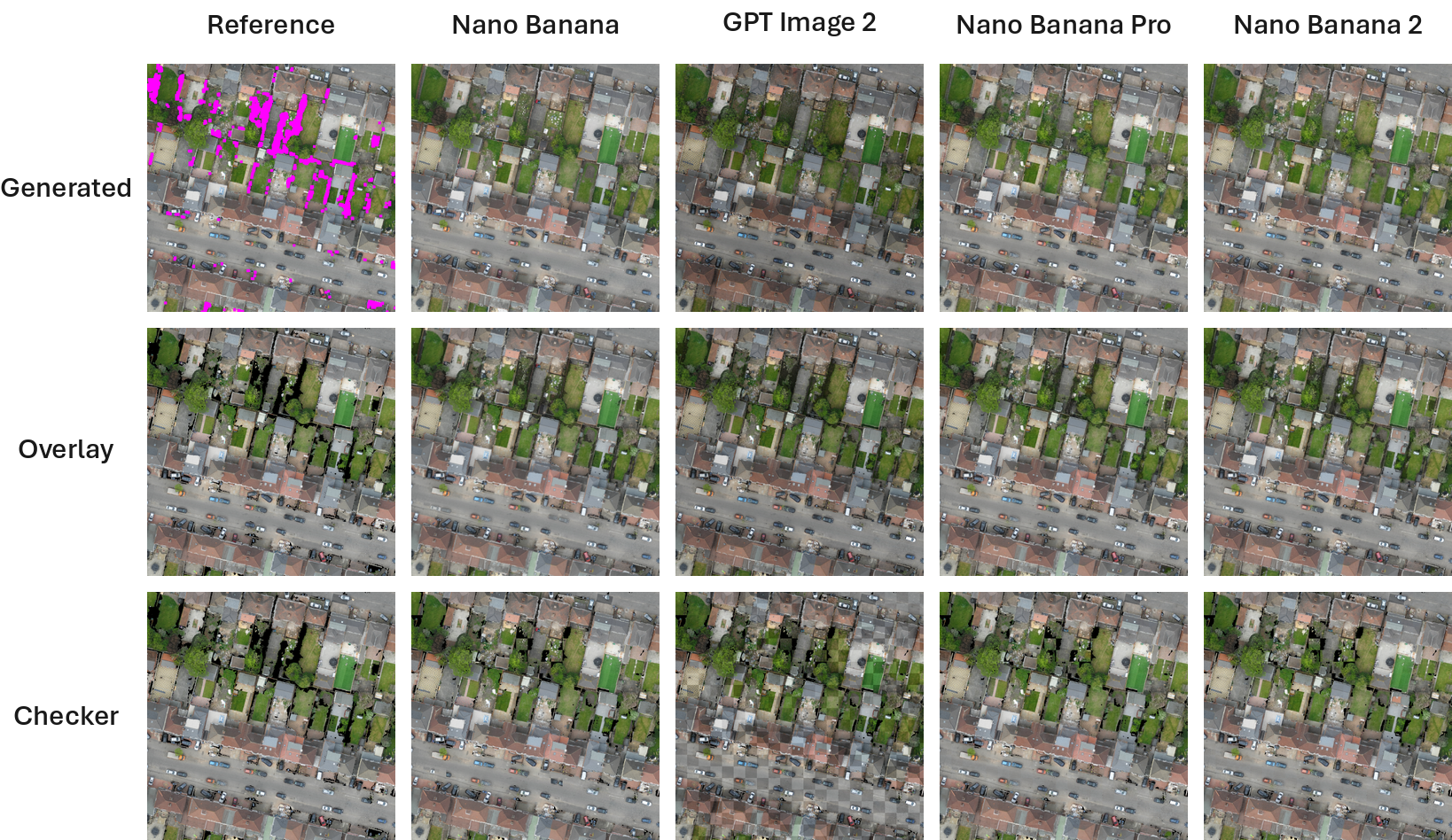}
    \caption{Comparison of RGB inpainting backbones on a large land-region hole. The larger missing area tests long-range texture continuation and urban-structure recovery. GPT Image 2 exhibits a systematic color shift, while Nano Banana 2 achieves a more consistent transition between the generated content and the surrounding valid image.}
    \label{fig:app_rgb_model_large_land}
\end{figure}

\begin{figure}[!t]
    \centering
    \includegraphics[width=\textwidth]{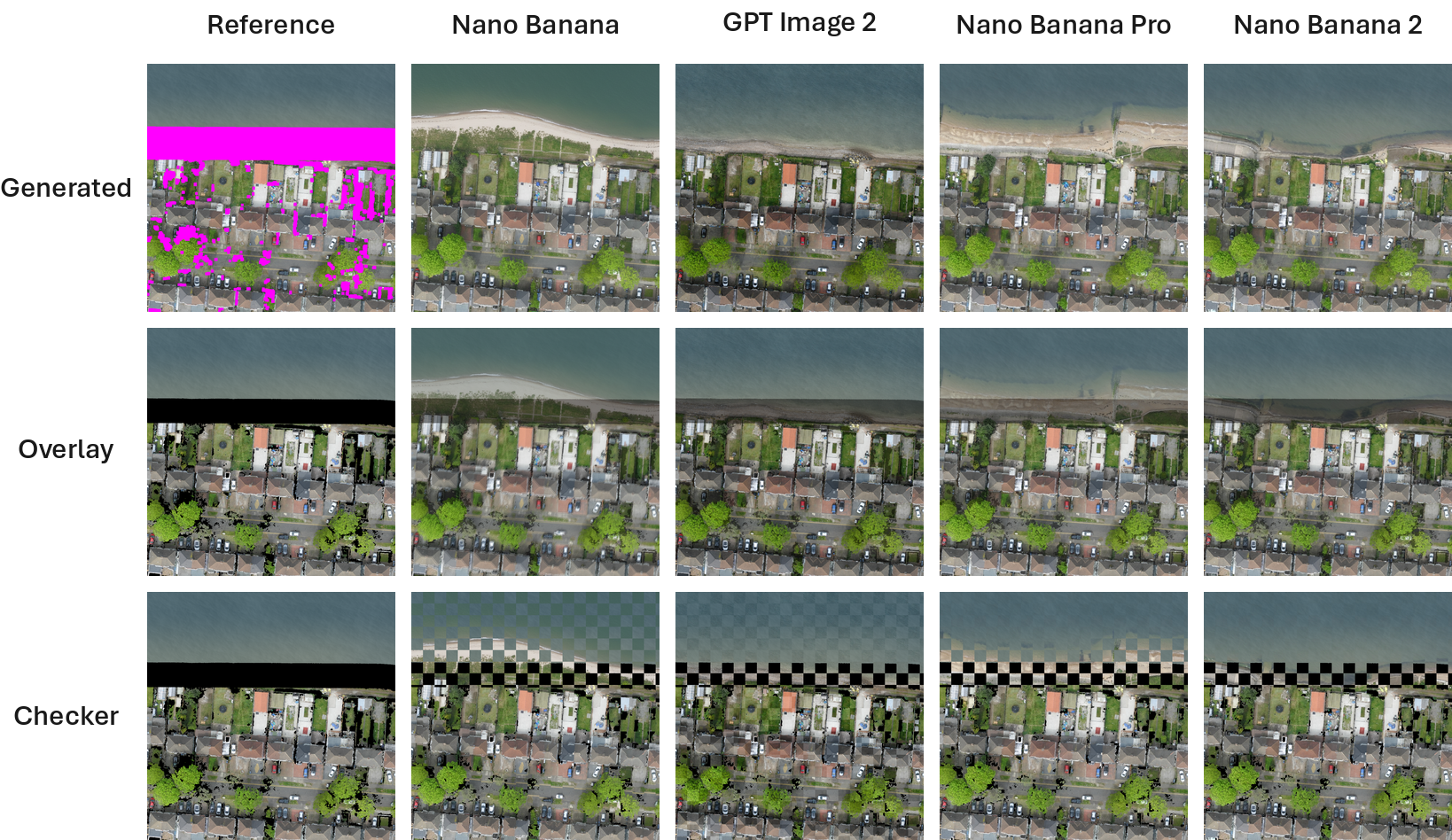}
    \caption{Comparison of RGB inpainting backbones near a coastline. Nano Banana and GPT Image 2 modify content beyond the prescribed mask boundary, producing unsafe writeback regions. Nano Banana 2 better respects the target mask while maintaining the land--water boundary.}
    \label{fig:app_rgb_model_coastline}
\end{figure}

\begin{figure}[!t]
    \centering
    \includegraphics[width=\textwidth]{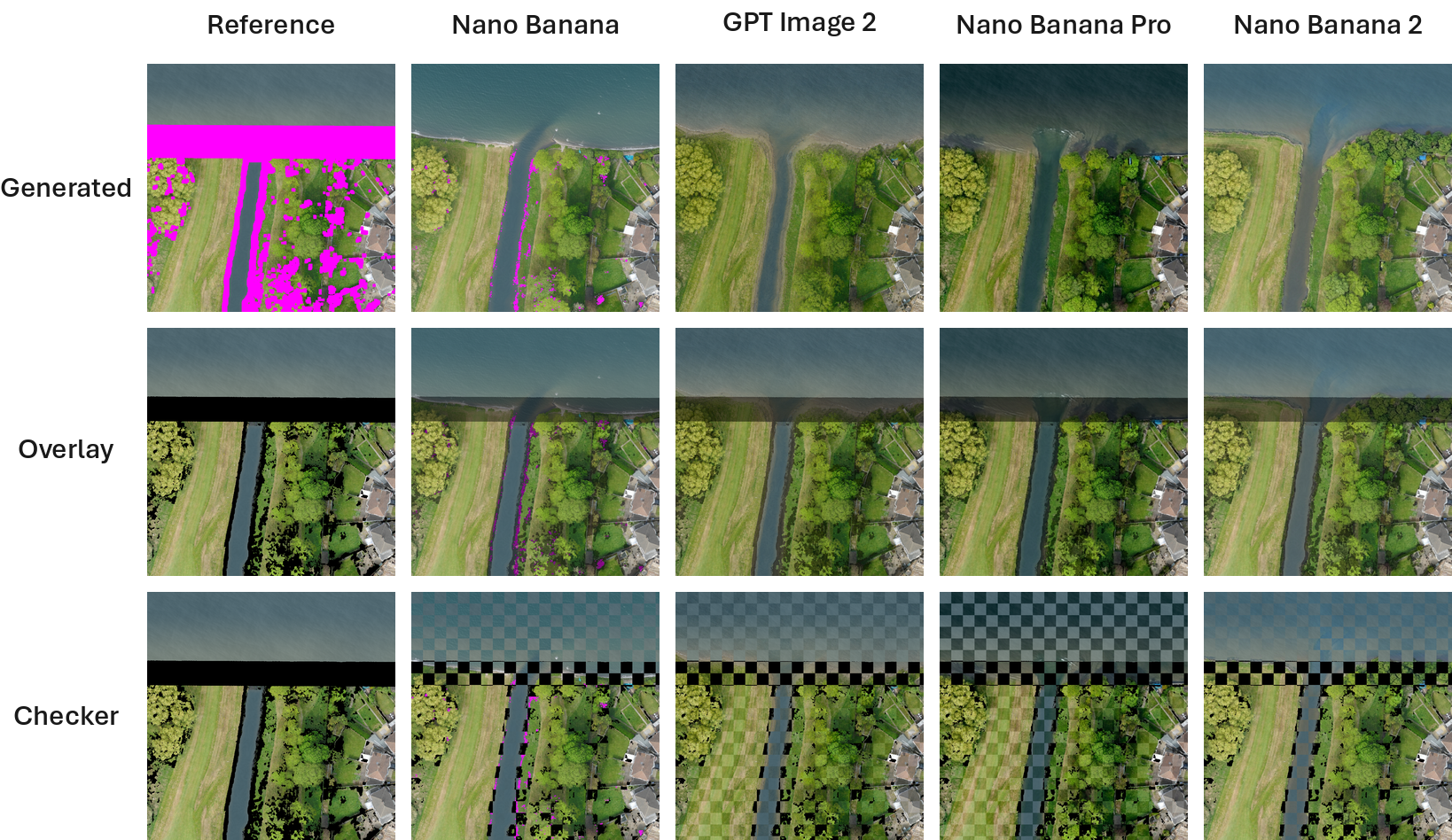}
    \caption{Comparison of RGB inpainting backbones on an estuary region. Nano Banana leaves residual masked pixels, while GPT Image 2 introduces the same global color shift observed in the land cases. Nano Banana 2 fills the target region more completely, although this difficult water-boundary case can still exhibit mild tone drift.}
    \label{fig:app_rgb_model_estuary}
\end{figure}

GPT Image 2 consistently introduces global color drift relative to the surrounding valid context. Nano Banana often preserves local texture well, but is less reliable in geometric registration and complete mask removal. Nano Banana Pro improves some local completions but still exhibits unstable writeback behavior in difficult boundary cases. Nano Banana 2 gives the best overall balance among mask compliance, structural continuity, and color consistency, and is therefore used as the default RGB backend. Figs.~\ref{fig:app_rgb_model_land_alignment}--\ref{fig:app_rgb_model_estuary} show representative land, coastline, and estuary cases.

\subsection{Reviewer Filtering}

Reviewer filtering mainly removes three classes of failures. First, it rejects generations that modify regions outside the prescribed target mask. Second, it removes candidates with residual masked pixels or incomplete filling. Third, it catches water-boundary failures, including river tone drift, shoreline discontinuity, and mask-shaped artificial coastlines.

\begin{figure}[!t]
    \centering
    \includegraphics[width=\textwidth]{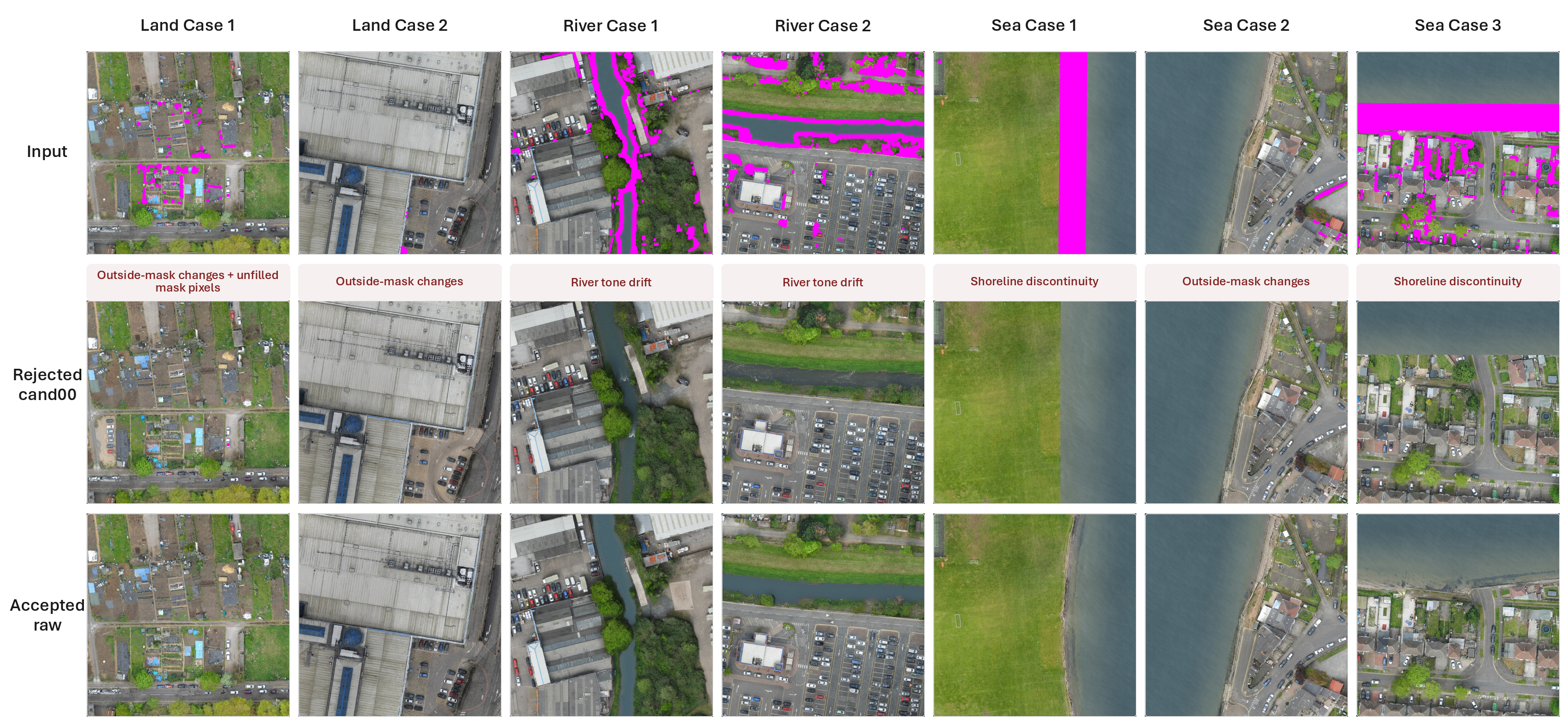}
    \caption{Examples of reviewer-based candidate filtering across land, river, and sea regions. Each column shows a masked input, the rejected first candidate with its principal failure highlighted, and the final accepted generation. The reviewer removes mask leakage, residual unfilled pixels, urban-structure changes, river tone drift, and broken shoreline continuity before global writeback.}
    \label{fig:app_rgb_reviewer_filtering}
\end{figure}

This filtering is important because many incorrect generations are locally plausible when viewed as isolated crops but become visible artifacts after global writeback. The reviewer therefore acts as a consistency gate between local image generation and city-scale canvas composition. Representative rejected and accepted candidates are shown in Fig.~\ref{fig:app_rgb_reviewer_filtering}.

\subsection{Overlap and Writeback}

We use overlapping sliding windows rather than hard non-overlapping tiling. In a hard \(2 \times 2\) tiling baseline, the same \(2048 \times 2048\) crop is split into four independent \(1024 \times 1024\) tiles, each completed and stitched directly. Even when the four local outputs are individually plausible, direct stitching introduces visible seams along internal tile boundaries, especially when the invalid region crosses the middle horizontal or vertical split.

\begin{figure}[!t]
    \centering
    \includegraphics[width=\textwidth]{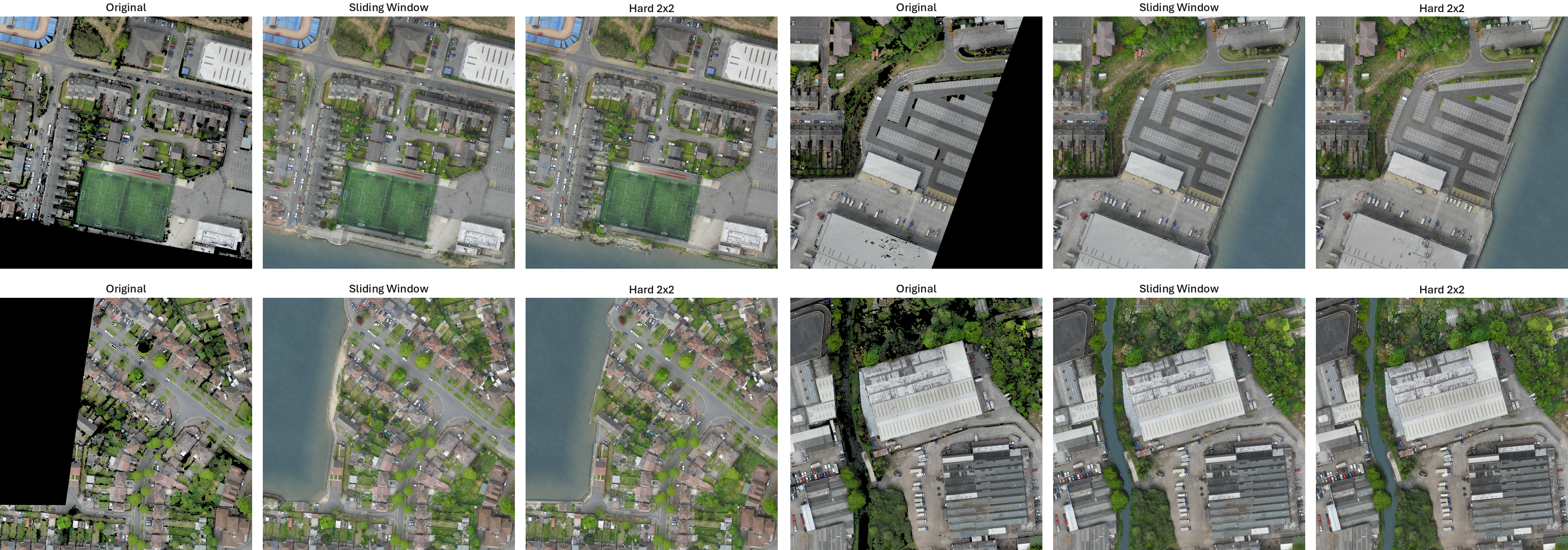}
    \caption{Comparison between overlapping-window completion and hard non-overlapping tiling on large Birmingham crops. Columns show the incomplete input, the result of the proposed overlapping sliding-window pipeline, and the hard $2 \times 2$ stitching baseline. Independent tiles create visible horizontal and vertical seams when a missing region crosses an internal boundary, whereas shared context and center-only writeback substantially improve global continuity.}
    \label{fig:app_rgb_overlap_stitching}
\end{figure}

The overlapping planner gives adjacent windows shared context and writes back only the more reliable center region. As shown in Fig.~\ref{fig:app_rgb_overlap_stitching}, this substantially reduces seam artifacts and makes the procedure practical for native-resolution city-scale completion.

\subsection{Ablation}

The RGB ablations focus on typed prompting and reviewer-based filtering. We compare variants without tile typing, with prompt typing only, with reviewer typing only, and with the full typed RGB pipeline. Prompt typing mainly improves local semantic plausibility, while reviewer typing rejects candidates that remain locally plausible but break global consistency after writeback. The full typed agent is necessary to simultaneously preserve region-specific appearance and suppress large-canvas seam artifacts.

\end{document}